\documentclass{article}

\PassOptionsToPackage{numbers, compress}{natbib}
\usepackage[final]{neurips_2023}

\usepackage[utf8]{inputenc} 
\usepackage[T1]{fontenc}    
\usepackage{hyperref}       
\usepackage{url}            
\usepackage{booktabs}       
\usepackage{amsfonts}       
\usepackage{nicefrac}       
\usepackage{microtype}      
\usepackage{xcolor}         

\usepackage{amsmath}
\usepackage{algpseudocode}
\usepackage{xcolor}
\usepackage{multirow}
\usepackage{array}
\usepackage{graphicx}
\usepackage{booktabs}
\usepackage{enumitem}
\usepackage{epsfig}
\usepackage{graphicx}
\usepackage{soul}
\usepackage{booktabs}
\usepackage{wrapfig}
\usepackage{footnote}
\usepackage{subcaption}
\usepackage{booktabs,tabularx}
\usepackage{cleveref}
\usepackage[ruled]{algorithm2e} 

\usepackage[T1]{fontenc}
\usepackage[utf8]{inputenc}
\usepackage{babel}
\usepackage[font=small,labelfont=bf]{caption}
\usepackage{graphicx}

\title{CLIPtortionist: Zero-shot Text-driven Deformation for Manufactured 3D Shapes}

\author{%
  Xianghao Xu \\
  Brown University, USA\\
  \text{xianghao\_xu@brown.edu} \\
  \And
  Srinath Sridhar\\
  Brown University, USA\\
  \text{srinath\_sridhar@brown.edu} 
  \And
  Daniel Ritchie\\
  Brown University, USA\\ 
  \text{daniel\_ritchie@brown.edu}
}

\begin{document}

\maketitle

\begin{center}
    \centering
    \captionsetup{type=figure}\includegraphics[width=0.98\linewidth]{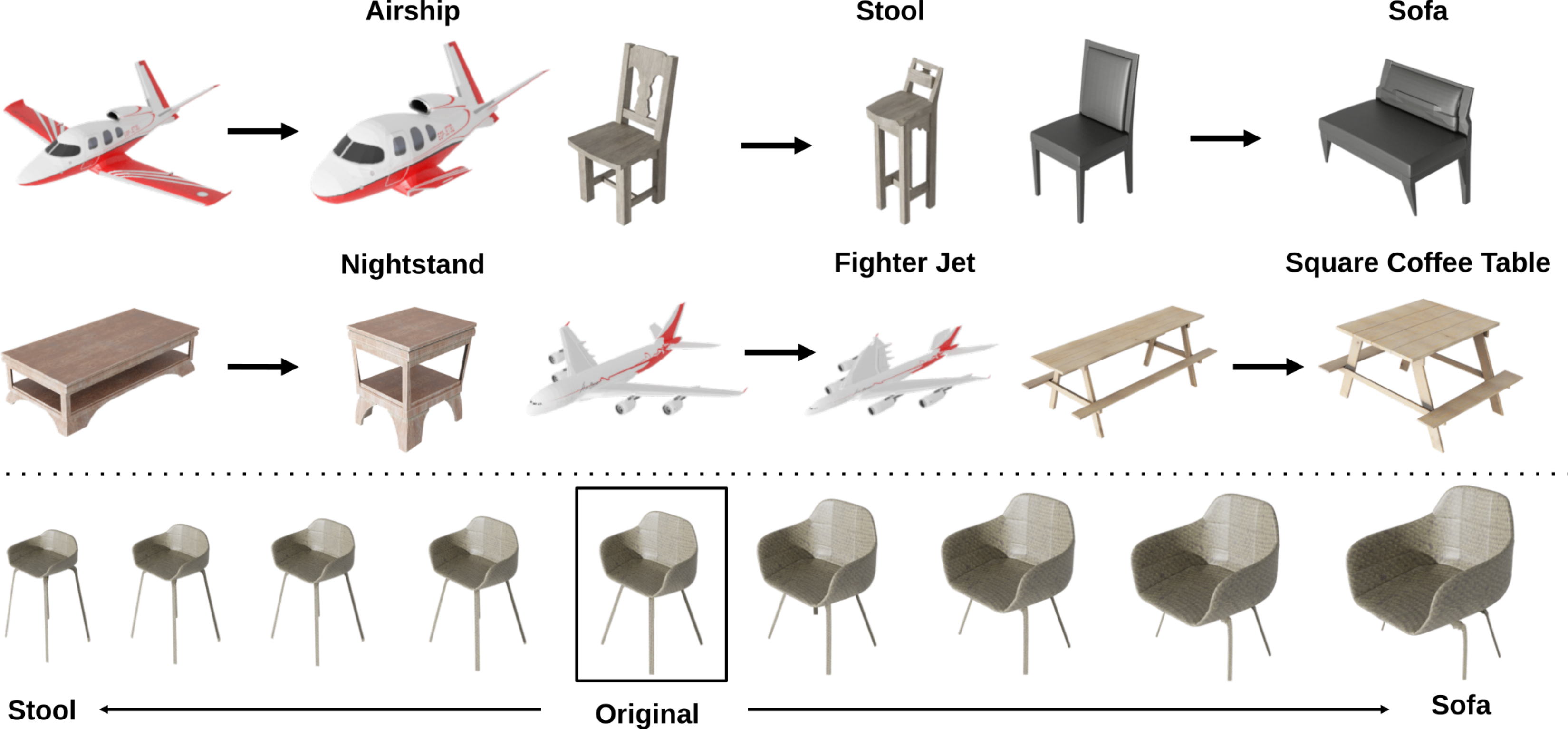}
    \captionof{figure}{We present CLIPtortionist, a zero-shot text-driven 3D shape deformation method that can deform an input 3D shape according to a text description while preserving structure and geometry.
    (Bottom)~Our method deforms shapes smoothly from a source shape (middle) to two targets specified using only text.
    }
    \label{fig:teaser}
\end{center}%

\begin{abstract}
We propose a zero-shot text-driven 3D shape deformation system that deforms an input 3D mesh of a manufactured object to fit an input text description. To do this, our system optimizes the parameters of a deformation model to maximize an objective function based on the widely used pre-trained vision language model CLIP. We find that CLIP-based objective functions exhibit many spurious local optima; to circumvent them, we parameterize deformations using a novel deformation model called BoxDefGraph which our system automatically computes from an input mesh, the BoxDefGraph is designed to capture the object aligned rectangular/circular geometry features of most manufactured objects. We then use the CMA-ES global optimization algorithm to maximize our objective, which we find to work better than popular gradient-based optimizers. We demonstrate that our approach produces appealing results and outperforms several baselines. 
\end{abstract}

\section{Introduction}
The ability to edit manufactured 3D shapes (for instance, airplane, chair, or table) is integral to many application domains, including games and industrial product design. In these applications, it is desirable to add, remove, or deform parts of the shape so as to reflect users' desire to change the geometry and semantic properties (e.g.,~tall chair, or bar stool) of the shape, often represented as a mesh. Furthermore, users can provide their input through the manipulation of a graphical user interface, or perhaps using a natural language interface. Deformation-based editing is a common form of shape editing, particularly for meshes (e.g.,~in 3D modeling software). One desired property users may want is to deform the bulk shape of the object without altering surface details (e.g.,~transforming an airliner to a fighter jet while preserving texture and other surface geometry details). 
Meshes can be edited by directly manipulating deformation handles such as vertices/faces, or some kind of proxy deformation handles with fewer degrees of freedom (e.g.,~a bounding cage). But it can also be desirable to provide \emph{higher-level} controls for editing that requires less precise user manipulation. For example, some prior work takes in a source shape and a target shape, then optimizes a deformation that warps the source mesh into a similar shape as the target mesh~\cite{Yifan:NeuralCage:2020}. However, the user may not always have access to such ``target'' shapes, particularly if they are working with a category of shape that is not well-represented in existing large shape repositories.

An alternative when target shapes are unavailable is to use a text description of the target shape.
This input can be easier to provide while capturing higher-level information about the target.
Prior work has tried to operationalize this idea: Semantic Shape Editing~\cite{semantic_shape} can produce large semantic deformations of the input shape in response to a text prompt, but it requires significant human supervision in the form of (1) manually-designed part-level deformers around the shape and (2) a large set of perceptual judgments from which the deformation optimization objective is derived.
Text2Mesh~\cite{text2mesh}, a more recent method, avoids the need to collect perceptual judgment data by instead defining an optimization objective based on the CLIP vision-language model~\cite{clip2021}. However, this method can only modify local surface details of the mesh and cannot produce large deformations that change the overall shape. Another recent method, TextDeformer~\cite{Gao_2023_SIGGRAPH}, deforms a mesh by optimizing the Jacobian of mesh faces and also uses CLIP as supervision. This method can make large deformations of a mesh, but it also changes the surface details and sometimes even the structure of the mesh thus destroying its original identity.

We present \textbf{CLIPtortionist}, a text-driven shape deformation method that addresses the limitations of these prior methods. Like Text2Mesh and TextDeformer, our method uses the pre-trained CLIP~\cite{clip2021} model to guide the deformation toward outputs that are semantically close to the text prompt. Like Semantic Shape Editing, we are able to produce large deformations of the input shape to satisfy a text prompt and preserve the texture and fine surface details. Achieving these two properties simultaneously is challenging: using CLIP as an optimization objective produces many spurious local optima in which naive deformation parameterizations are likely to fail. To solve this problem, we parameterize our mesh deformations using a novel mesh deformer, called BoxDefGraph, which we designed to avoid the worst of these local optima. BoxDefGraph is a part-level Object Aligned Bounding Box (AABB) graph that is automatically constructed from an input shape by analyzing the shape's geometry properties and deformation constraints. Our deformation model can make large deformations to the input mesh to better fit the text prompt while maintaining the mesh's fine details. This deformation model is specifically designed for manufactured objects as these objects tend to have well-defined rectangular/circular geometry features that are aligned with the object bounding boxes. So that a set of object-aligned bounding boxes with each of them capturing the geometry of a local region can capture most manufactured objects. Another advantage of our deformation model is that by interpolating the parameters associated with the deformers, a continuous spectrum of deformations toward (or away from) the concept expressed by the text prompt can be obtained (see \Cref{fig:teaser}). Even with a well-parameterized deformation model, it is non-trivial to achieve large meaningful deformations by directly using gradient-based optimization in CLIP space because of its local optima. We avoid this problem by adopting a gradient-free, population-based global optimization method: Covariance Matrix Adaptation Evolutionary Strategy (CMA-ES) to optimize the deformation parameters. CMA-ES is only feasible in our setting because the carefully-designed BoxDefGraph deformation model has sufficiently few degrees of freedom. 

With the designed deformation model and optimization setting, our system can produce meaningfully large deformation to the input mesh to fit the text while still maintaining fine details of the input mesh. 
To summarize, our contributions are:
\begin{enumerate}[topsep=0pt,itemsep=-1ex,partopsep=1ex,parsep=1ex]
\item A zero-shot text-driven 3D shape deformation framework that can produce large semantic shape deformations to manufactured objects.
\item The deformation model BoxDefGraph, which imposes constraints between part level AABB deformers for 3D shape deformation.
\end{enumerate}

\section{Overview}

\begin{figure}[t!]
    \centering
    \includegraphics[width=0.9\linewidth]{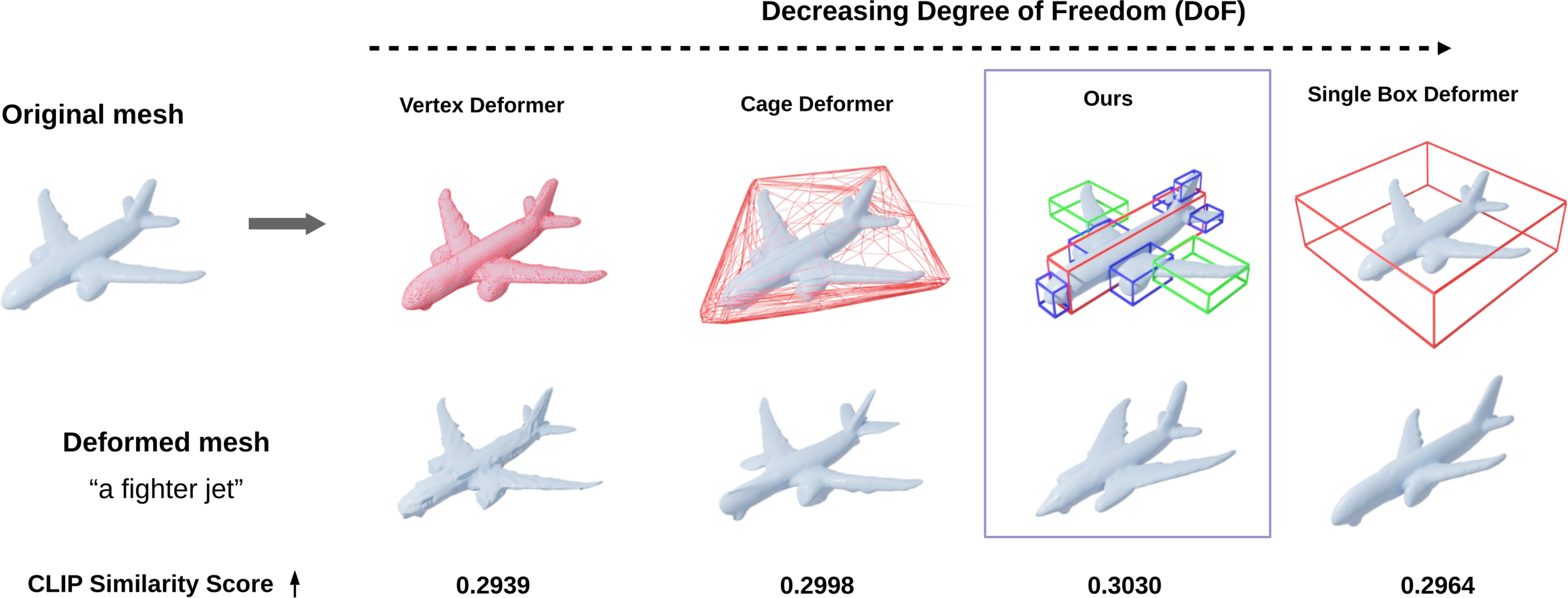}
    \caption{
Meshes deformed by different deformation models to maximize a CLIP-based similarity socre. From left to right, in decreasing order of degrees of freedom (DoF): Vertex deformer, Cage deformer, BoxDefGraph (Ours), and Single Box deformer. Our BoxDefGraph deformer's output matches the text prompt better than the other deformers.
}
    \label{fig:overview}
\end{figure}

The goal our CLIPtortionist{} system is to take a 3D triangular mesh $M$ and a short text prompt $T$ as input and output the deformed mesh $M_T$ that fits the description of the text prompt. The key to solving the problem lies in figuring out how to parameterize and optimize a mesh deformer. This is a non-trivial problem. The deformer should be able to introduce large deformation to the mesh while still maintaining its fine geometry details; furthermore, the only supervision signal it has access to is the input text prompt $T$. Not all mesh deformer parameterizations are equally well-suited for this task. Figure~\ref{fig:overview} compares the behavior of four different deformers whose parameters are optimized to maximize similarity of the output shape to the text prompt ``a fighter jet.''
The simplest possible deformation model is to treat every vertex of the mesh as an optimizable degree of freedom.
However, this approach leads to a noisy result: surface details of the original mesh are lost, and the bulk shape of the mesh does not meaningfully change.
A lower-degree-of-freedom approach is to compute the mesh's convex hull and use it as a control cage for mean-value-coordinate-based mesh deformation which is a variation of ~\cite{mvc_cage} and \cite{Yifan:NeuralCage:2020}, While this approach can significantly change the mesh's bulk shape, the output mesh still exhibits undesirable distortion.
A face-level deformer (Jacobian of mesh faces) such as ~\cite{Gao_2023_SIGGRAPH} has a similar problem as the cage-based deformer. This distortion can be eliminated by using the mesh's bounding box as a deformation cage.
However, the drastically-reduced DoF of this deformer is insufficient to produce output shapes that match the text prompt.

\begin{figure*}[t!]
    \centering
    \includegraphics[width=1.0\linewidth]{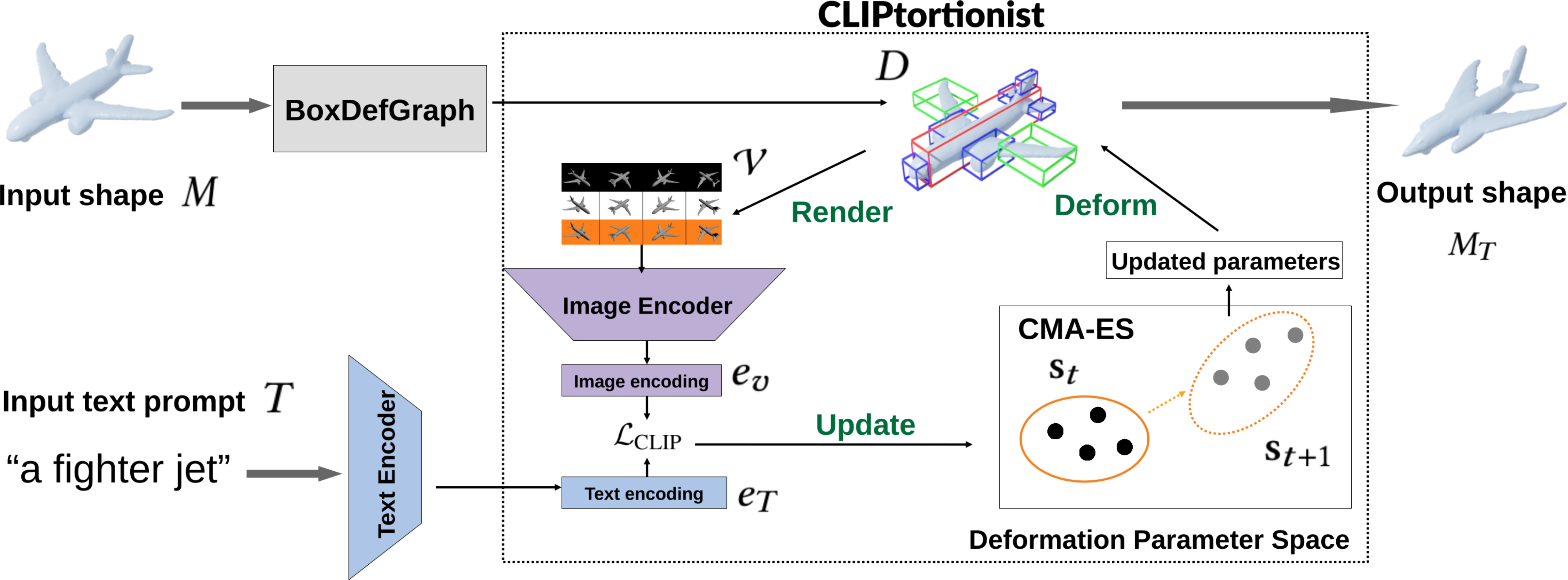}
    \caption{CLIPtortionist{} takes a 3D triangular mesh $M$ and a text prompt $T$ as input and outputs a deformed target mesh $M_T$ that fits the description in the text prompt. It first generates the deformation model BoxDefGraph $D$ by analyzing the input mesh $M$. Then the parameters of the deformation model BoxDefGraph are optimized in an iterative loop. The optimization loop starts by deforming the mesh according to the BoxDefGraph parameters $\mathbf{s}_t$. Then the deformed mesh is rendered into a collection of images $\mathcal{V}$, and each image $v$ is encoded by a pre-trained CLIP image encoder into a latent code $\mathbf{e}_v$. The input text $T$ is also encoded by a pre-trained CLIP text encoder into a latent code $\mathbf{e}_T$. The system then uses the CLIP Loss $\mathcal{L}_{\text{CLIP}}$ which is computed as the negative average cosine similarity of $e_v \in \mathcal{V}$ and $\mathbf{e}_T$. The parameters are updated by CMA-ES and the updated parameters $\mathbf{s}_{t+1}$ are used to deform the mesh in the next iteration. Once the loop terminates, the optimal parameters are applied to produce the final output deformed mesh $M_T$.
    }
    \label{fig:framework}
\end{figure*}

In this paper, we demonstrate that part-level bounding boxes are an appropriate choice for deformation model for this task: they can make large geometry changes while maintaining the fine details of the original mesh. We designed a deformation model called BoxDefGraph: a graph structure whose nodes are part-level AABBs and whose edges denote parametric constraints between AABBs. We show how to construct a BoxDefGraph automatically from an input mesh (without any part-level structure annotations).
With the deformation model settled, we need to optimize the deformer's parameters with respect to some measure of how well the deformed mesh fits the given text prompt. We can leverage any pre-trained vision-language model for this task; in our experiments, we use CLIP ~\cite{clip2021}.
In prior work on zero-shot learning and inference with CLIP, gradient-based optimization methods are predominantly used.
However, in our problem setting, we found that gradient-based methods can stuck at local optima in the CLIP space, even with a well-designed deformation model.
Instead, we turn to a CMA-ES ~\cite{hansen2019pycma}, a gradient-free global optimization method based on evolving a population of samples in the domain of the objective function.
Figure~\ref{fig:framework} shows an overview of the complete CLIPtortionist{} system.
Given an input mesh $M$ and text prompt $T$, it first computes a BoxDefGraph for $M$ and a CLIP space embedding for $T$.
The BoxDefGraph deformation model is then optimized in an iterative process.
Each iteration starts by deforming the mesh according to the BoxDefGraph's current parameters.
Then, the deformed mesh is rendered and embedding into CLIP space by a pre-trained CLIP image encoder, and this embedding is compared to the embedding of $T$ and used to guide deformation parameter optimization.
Once the optimization terminates, the optimal  BoxDefGraph parameters are applied to produce the final deformed mesh.

\section{Deformation Model}
\label{deformation_model}
We first describe our deformation model in detail before discussing how it enables text-driven deformation (\Cref{optimizing_for_deformations}). Our mesh deformer, called BoxDefGraph, is designed to enable large shape deformations with high-level text input.
In \Cref{BoxDefGraph}, we define our BoxDefGraph structure and how it applies deformations to a mesh. We then describe how to automatically construct a BoxDefGraph from an input mesh (\Cref{BoxDefGraph_construction}).
%

\subsection{BoxDefGraph}
\label{BoxDefGraph}

Our BoxDefGraph deformation model is central to our method and is formally defined as a graph $D = \{\mathcal{B}, \mathbf{C}\}$ with each node representing a part-level Axis-Aligned Bounding Box (AABB) $B_i \in \mathcal{B}$ and each edge representing the constraint $C_{ij} \in \mathcal{C}$ between the adjacent nodes. Each node AABB $B_i$ has parameters $\mathbf{s}_i \in R^3$ to scale the mesh vertices within it thus transforming the mesh locally. We limit the deformation parameters to only scaling for two main reasons: (1) to keep the number of optimizable parameters lower (2) To make it easier to keep the structure of most manufactured objects. Although it is straightforward to add translation and rotation. Note that it is insufficient to deform each AABB independently -- we also need to preserve the global geometry feature and the global structure feature of the input mesh to some extent. For geometry preservation, we achieve this by enforcing a vertex normal consistency loss. For structure preservation, we designed edge-level operations to preserve the relative position of vertices in different AABBs to some extent.
This structure preservation will also serve to preserve continuous features such as textures on the surface of the mesh. To formalize and satisfy these preservation constraints, we could turn them into terms in an objective function and apply optimization techniques. However, the set of box parameter configurations that satisfy constraints is small compared to the set of all possible box parameter configurations, so the optimizer may take a long time to converge (if it converges at all). Thus, we instead seek a reparameterization that allows as many of the constraints as possible to be satisfied by construction. We define a tree structure BoxDefTree as $D_T = \{\mathcal{B}, \mathbf{C_T}\}$ with nodes are nodes in graph $D$ and edges are a subset of edges in $D$. With this tree structure, the majority of the constraints can be actively satisfied by construction with executing appropriate tree-edge constraint-solving operations along the tree edges. And the leftover constraints will be handled by graph-edge constraint-solving operations. In the following paragraphs, we will describe how each constraint is defined and satisfied.

\subsubsection{Geometry Preservation}

The geometry preservation is achieved by enforcing a normal consistency term added to the optimization objective. The normal consistency term allows the deformer enough freedom to make large deformations of the shape of an object while still preserving the geometry features. For example, an-isotropic scaling of a flat region in the shape won't increase the normal consistency penalty value. The normal consistency penalty term is formulated as the cosine similarity between vertex normals $\mathbf{N}_v^{\text{ori}}$ before the deformation and vertex normals $\mathbf{N}_v^{\text{def}}$ after the deformation. $\mathcal{L}_{\text{normal}} \gets \frac{\mathbf{N}_v^{\text{ori}} \cdot \mathbf{N}_v^{\text{def}}}{|\mathcal{N}_v|}, \textbf{N}_v \in \mathcal{N}_v$
The vertex normal is computed by averaging the normals of the neighboring faces referring to the vertex $\mathbf{N}_v \gets \frac{\sum \mathbf{N}_f}{|\mathcal{F}|} , f \in \mathcal{F}$.

\subsubsection{Structure Preservation}
\label{connectivity_constraint}

Structure preservation is another crucial component. The key idea is trying to maintain the relative positions of vertices in different box deformer stay similar before and after the deformation. We handle this constraint in two steps: Box structure preservation and Boundary vertex structure preservation.

\paragraph{Box structure preservation} This constraint tries to maintain the relative position of boxes before and after the deformation. Taking an airplane as an example: when the body AABB of the airplane shrinks, all the mesh vertices in the AABB wrapping the airplane's tail region should move forward as a whole to make the tail stay close to the deformed airplane body. One thing to notice is that if we apply this constraint to all graph edges, then it will impose too much constraints and make the deformation impossible in some cases, for example, deforming a table top surface with two legs initially contacted, if we want to enlarge the surface and keep two legs each on one side of this enlarged surface, then the connectivity between two legs have to be cut. So instead, we apply this first structure preservation constraint to only the tree edges. Formally, we define:
\begin{equation*}
\begin{aligned}
\mathbf{p}_i, \mathbf{p}_j &= \text{argmin}_{\mathbf{p}_i^{'}, \mathbf{p}_j^{'}}\{f = ||\mathbf{p}_i^{'} -\mathbf{p}_j^{'}||_2, \mathbf{p}_i^{'} \in \mathbf{P}_i^{s}, \mathbf{p}_j^{'} \in \mathbf{P}_j^{f} \},\\
\mathbf{r}_{ij} &= \mathbf{p}_i - \mathbf{p}_j,\\
\mathbf{\text{TreeEdgeOp}}_{ij} &= \mathbf{r}_{ij}^{\text{def}} - \mathbf{r}_{ij}^{\text{ori}},
\end{aligned}
\end{equation*}

where for each tree edge with parent node $B_i$ and child node $B_j$, we define $\mathbf{r}_{ij}$ as the relative position vector between $B_i$ and $B_j$ computed as the vector connecting two points $\mathbf{p}_i$ and $\mathbf{p}_j$, the shortest distance between two point sets $\mathbf{P}_i^{s}$ and $\mathbf{P}_j^{f}$.
$\mathbf{P}_i^{s}$ contains the equally-spaced sampled surface points of $B_i$, $\mathbf{P}_j^{f}$ contains the face centers of $B_j$.
The operation $\mathbf{\text{TreeEdgeOp}}_{ij}$ satisfies this first connectivity constraint by translating the mesh vertices $\mathbf{V}_j$ in $B_j$ by the difference vector between the new relative position vector $\mathbf{r}_{ij}^{\text{def}}$ after the deformation and the original relative position vector $\mathbf{r}_{ij}^{\text{ori}}$.\\

\paragraph{Boundary vertex structure preservation} This constraint tries to maintain the relative positions of vertices around box boundaries before and after the deformation. Taking an airplane as an example: the mesh vertices within the airplane body AABB may scale vertically and the mesh vertices within the airplane tail AABB may scale horizontally, thus introducing sharp changes of mesh vertex positions around the contact region. This constraint is applied to all graph edges.
Formally, we define
\begin{equation*}
\begin{aligned}
\mathbf{V}_i^{b} &= \bigcup_{\forall \mathbf{p}_j^{'}} \{{\text{argmin}_{\mathbf{p}_i^{'}}\{f = ||\mathbf{p}_i^{'} - \mathbf{p}_j^{'}||_2, \mathbf{p}_i^{'} \in \mathbf{V}_i, \mathbf{p}_j^{'} \in \mathbf{V}_j \}}\},\\
\mathbf{F}_{ij} &= \mathbf{V}_i^{b} - \mathbf{V}_j,\\
\mathbf{\text{GraphEdgeOp}}_{ij} &= 0.5 * \text{min}(1.0, \epsilon/||\mathbf{F}_{ij}^{\text{ori}}||_2) (\mathbf{F}_{ij}^{\text{def}}- \mathbf{F}_{ij}^{\text{ori}}),
\end{aligned}
\end{equation*}
where for each graph edge with node $B_i$ and node $B_j$, we define a vector field $\mathbf{F}_{ij}$ which is computed as the position difference vectors between mesh vertices $\mathbf{V}_j$ in $B_j$ to their nearest mesh vertices $\mathbf{V}_i^{b}$ in $B_i$. The operation $\mathbf{\text{GraphEdgeOp}}_{ij}$ to satisfy the constraint is to translate the mesh vertices $\mathbf{V}_j$ in the node $B_j$ by the vector field difference between the new vector field $\mathbf{F}_{ij}^{\text{def}}$ after the deformation and the original vector field $\mathbf{F}_{ij}^{\text{ori}}$, multiplied with the per-vertex weights as a decaying factor vector according to the original vector field $\mathbf{F}_{ij}^{\text{ori}}$. The symmetrical operation is also performed with the role of node $B_i$ and $B_j$ swapped.

\subsubsection{Execution}
A deformation is executed by traversing the BoxDefTree of the BoxDefGraph from the root node to the leaves. The scaling parameters of each node's associated AABB are applied to the AABB and to every vertex within the AABB; then, the tree edge constraint operation between the parent node AABB and child node AABB is executed, and finally the graph edge constraint operation between each pair of adjacent AABBs is executed. Please see supplement for more details.




\subsection{BoxDefGraph Construction}
\label{BoxDefGraph_construction}
To build a BoxDefGraph from an input mesh, we first use a part-level AABB generation process which is a modified version of \cite{autoOBB}.
The idea is to start with the shape-level Axis-Aligned Bounding Box (AABB) and recursively split the upper-level boxes into lower-level axis-aligned bounding boxes that wrap each local region of the mesh until the desired split number is reached. In order to control the splitting degree, we assign a split priority (reciprocal of a \emph{split score} $s_c$) to each AABB.
At each iteration of the algorithm, the AABB $B_i$ with the highest split priority is cut into two child AABBs $B_{i}^{'}$ and $B_{i}^{''}$. To compute the split score $s_c$, we first rasterize the mesh onto a 3D occupancy grid; then, for each AABB $B_i$, a linear scan with resolution $\delta$ along each axis $a_j$ of $B_i$ is performed. At each potential cut section $c$, the 2D projection grid $g_c$ of the occupancy voxels in the section $vox_{c:c+\delta}$ is computed.
Then the split score $s_c$ is computed as the area change ratio between adjacent projection grids $g_c^{t}$ and $g_c^{t-1}$, multiplied by the reciprocal of the axis length $B_i.L_j$ of $B_i$.
The score is given as,
\begin{align}
s_c \gets \frac{\text{min}(\sum{g_c^t}, \sum{g_c^{t-1}})}{\text{max}(\sum{g_c^t}, \sum{g_c^{t-1}})} * \frac{1.0}{B_i.L_j}.\nonumber
\end{align}
The lower the score, the more it indicates a drastic geometry change in a longer axis, thus indicating a good split position.
Please see the supplement for further details of this AABB split procedure.
In our experiments, we iterate over split counts [2, 3, 4] and pick the result that gives the maximum objective function value. Once the AABBs are generated, the BoxDefGraph $D$ is naturally constructed, and the tree structure BoxDefTree $D_T$ is built by running a Breath-First-Search (BFS) over the generated AABBs $\mathcal{B}$. Starting with the largest AABB as the root AABB, the adjacent neighbor AABBs are retrieved by checking if there is any face edge of the mesh bridging the two AABBs. The parents nodes that car connected to the same child node will be merged. The constraints $C_{ij}$ between parent AABB $B_i$ and child AABB $B_j$ are precomputed during the tree construction for fast query later in the deformation process.  

\section{Deformation Optimization}
\label{optimizing_for_deformations}

\begin{algorithm}
\small
\DontPrintSemicolon
\SetKwInOut{Input}{Input}
\SetKwInOut{Output}{Output}
\Input{Input mesh $M$, Input text $T$, BoxDefGraph $D$, }
\Output{Deformed mesh $M_T$.}
\SetKwProg{Fn}{Function}{:}{}
\SetKwFunction{FOptim}{Optim}
\Fn{\FOptim{$M, T, D$}}{
$\mathbf{e}_{T} \gets \text{Normalize}(\text{TextEncoder}(T))$\;
$\mathbf{s}_0 \gets \text{Initialize()}$\;
\While {$t \le \text{maxIter}$} {
    $M_t \gets \text{Deform}(M, D, \mathbf{s}_t)$\;
    $\mathcal{V} \gets \text{Render}(M_t)$\;
    $\mathbf{e}_{v} \gets \text{Normalize}(\text{ImageEncoder}(v)), v \in \mathcal{V}$\;
    $s_v \gets \text{CosineSimilarity}(\mathbf{e}_{T}, \mathbf{e}_{v})), v \in \mathcal{V}$\;
    $\mathcal{L}_{\text{CLIP}} \gets -\text{Avg}(s_v), v \in \mathcal{V}$ \;
    $\mathcal{L} = \mathcal{L}_{\text{CLIP}} + \mathcal{L}_{\text{normal}}$\;
    $\mathbf{s}_{t+1} \gets \text{CMAES}.\text{Next}(\mathbf{s}_t, \mathcal{L})$ \;
}
$M_T \gets \text{Deform}(M, D, \mathbf{s}_{opt})$\;
\Return $M_T$\;
}
\caption{
    Deformation Model Optimization
    }
\label{alg:optim}
\end{algorithm}

With the deformation model and associated parameters defined, we describe how the deformation model is optimized to deform the mesh to fit the input text description. The main supervision that measures the fitness between the deformed mesh $M_t$ and input text $T$ comes from the CLIP similarity. The similarity is computed as follows: the deformed mesh $M_t$ at iteration $t$ is rendered into a set of equally spaced views $\mathcal{V}$ (The mesh is colored grey and the background is colored with three different colors: white, black and orange). The rendered image $v$ of each view is encoded into the CLIP latent space by the pre-trained CLIP image encoder as $\mathbf{e}_v$; correspondingly, the input text prompt $T$ is encoded to the same CLIP latent space by the pre-trained text encoder as $\mathbf{e}_T$. The cosine similarity between each view encoding vector $\mathbf{e}_v$ and the text encoding vector $\mathbf{e}_T$ is computed. The final CLIP similarity is obtained by averaging the similarities across all views. In addition to the fitness as the primary supervision signal, the normal consistency term $\mathcal{L}_{\text{normal}}$ mentioned in \Cref{deformation_model} can also be optionally added. The final loss function that will be minimized by the optimizer is defined as $\mathcal{L} = \mathcal{L}_{\text{CLIP}} + \mathcal{L}_{\text{normal}}$, where
\begin{equation*}
\begin{aligned}
\mathcal{L}_{\text{CLIP}} \gets -\frac{1}{|\mathcal{V}|}\sum_{v=1}^{|\mathcal{V}|}{\text{CosineSimilarity}(\textbf{e}_v, \textbf{e}_T)}
\end{aligned}
\end{equation*}
We use CMA-ES for the deformation model parameter optimization, because we found it was better at avoiding local optima and producing more consistent run-to-run outputs (which will be demonstrated in Section~\ref{results}). At each iteration, CMA-ES generates deformation parameter samples $\mathbf{s}_t$, evaluates them under $\mathcal{L}$, and uses the evaluated samples to generate the next iteration of deformation parameters $\mathbf{s}_{t+1}$. The optimization process terminates when it converges or reaches a specified maximum iteration number. Please see Algorithm~\ref{alg:optim} for more details.

\section{Results} \label{results}

We conducted several experiments with our deformation framework on Airplane, Chair and Table categories from the ShapeNet dataset ~\cite{shapenet2015}. 

\subsection{Comparison to Alternative Methods}
\begin{figure}[t!]
    \centering
    \small
    \setlength{\tabcolsep}{0pt}
    \renewcommand{\arraystretch}{0}
    \newcommand{\resultimg}[1]{\includegraphics[trim={60pt 60pt 60pt 60pt},clip,width=0.14\linewidth]{#1}}
    \begin{tabular}{lcccccc}
        & \raisebox{0em}{Fighter Jet} & \raisebox{0em}{Airship} & \raisebox{0em}{Stool} & \raisebox{0em}{Sofa} & \raisebox{0em}{Nightstand} & \raisebox{0em}{Square Coffee Table}
        \\
        \raisebox{2em}{Original} & 
        \resultimg{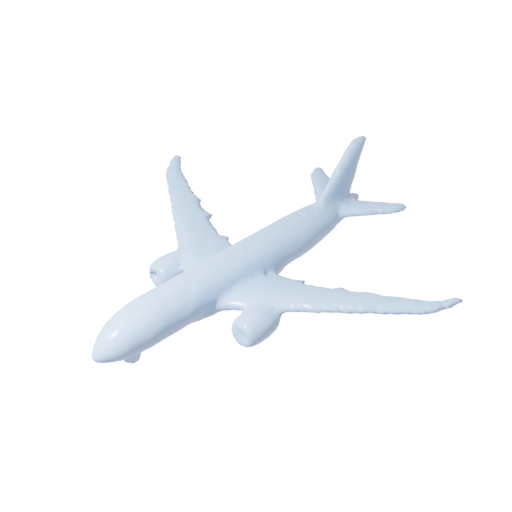} &
        \resultimg{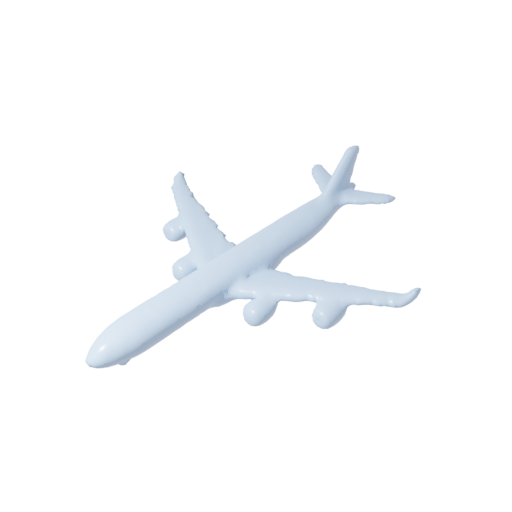} &
        \resultimg{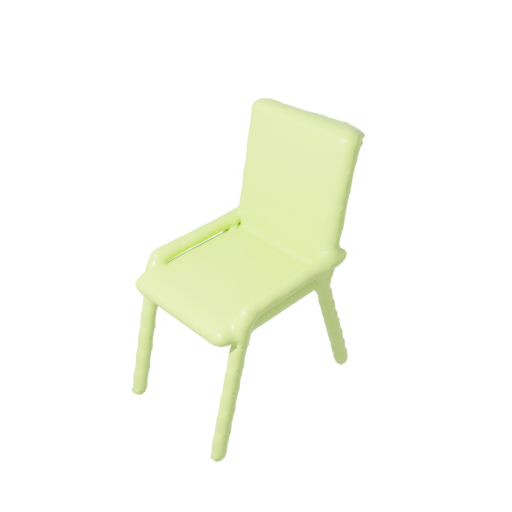} &
        \resultimg{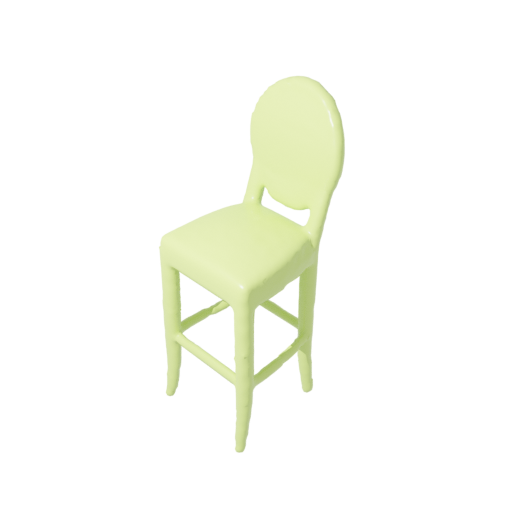} &
        \resultimg{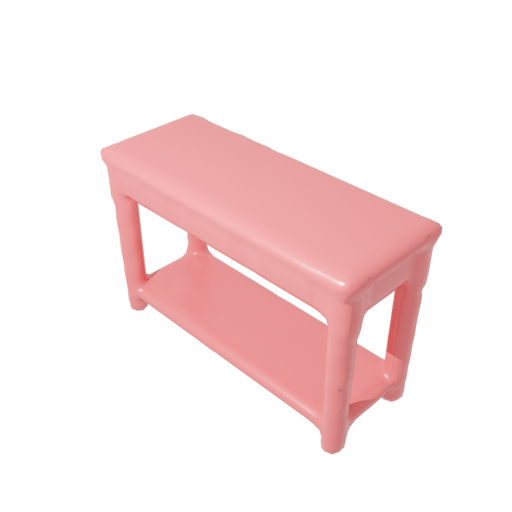} &
        \resultimg{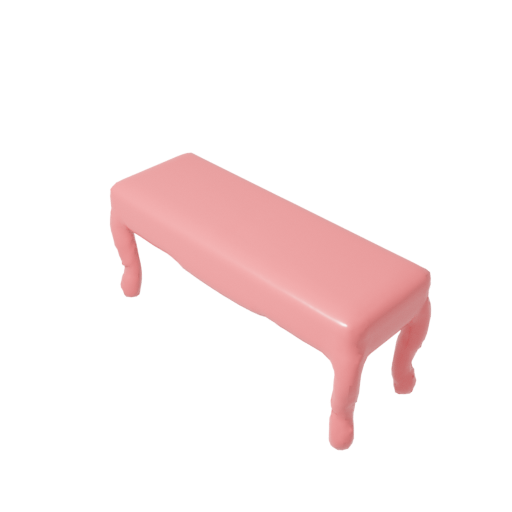} 
        \\
        \raisebox{2em}{Text2Mesh} & 
        \resultimg{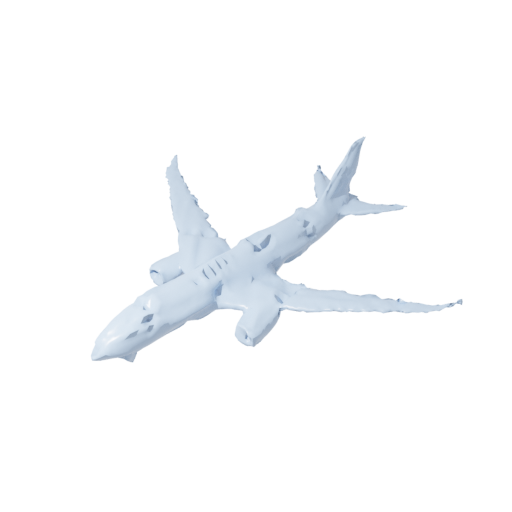} &
        \resultimg{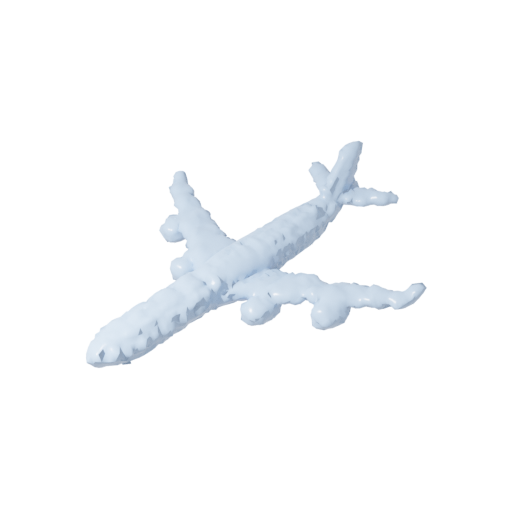} &
        \resultimg{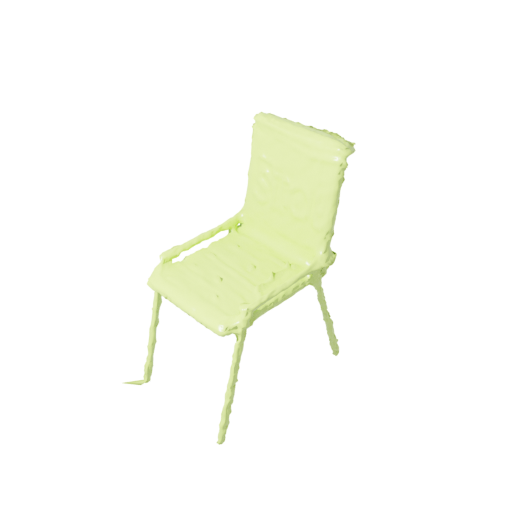} &
        \resultimg{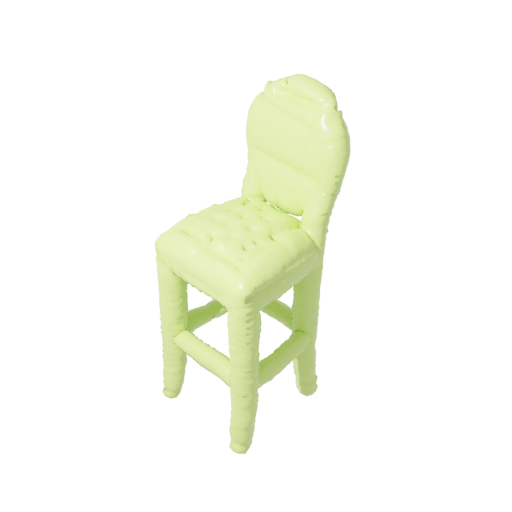} &
        \resultimg{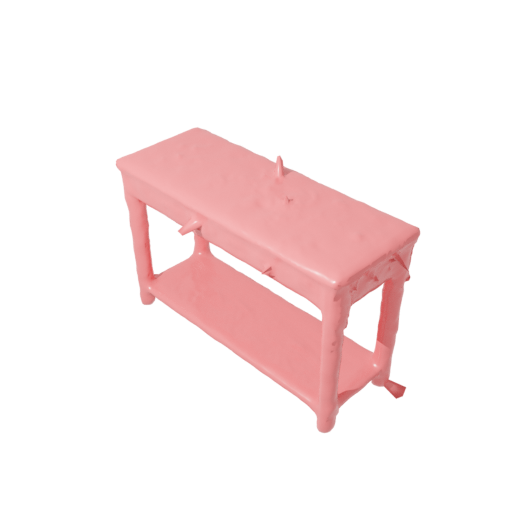} &
        \resultimg{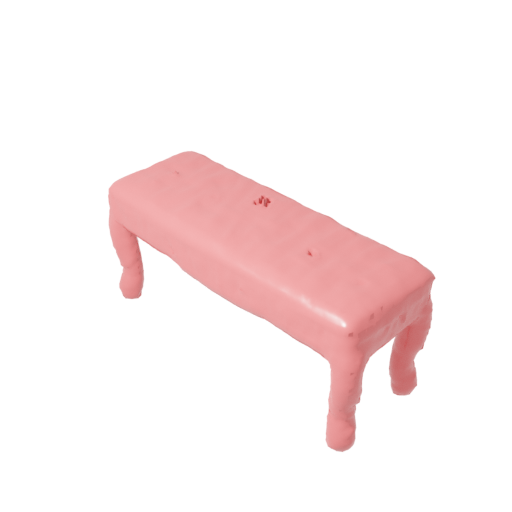} 
        \\
        \raisebox{2em}{TextDeformer} & 
        \resultimg{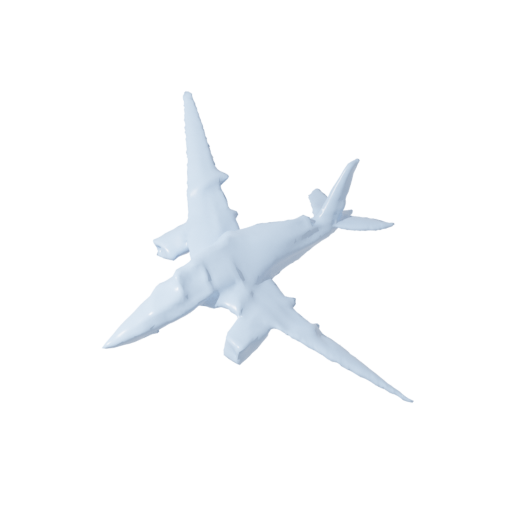} &
        \resultimg{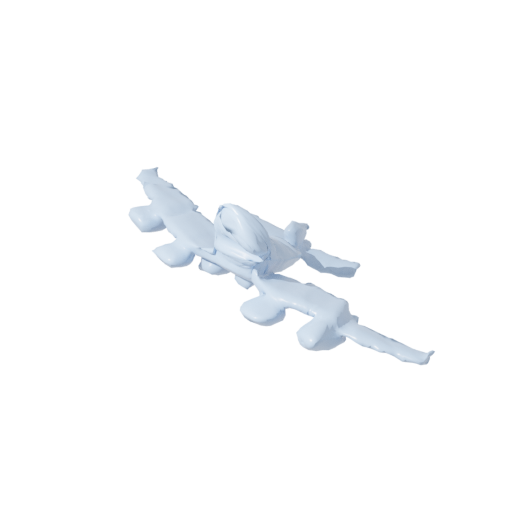} &
        \resultimg{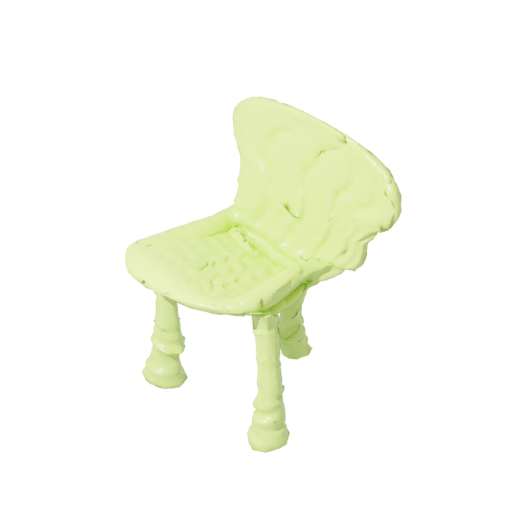} &
        \resultimg{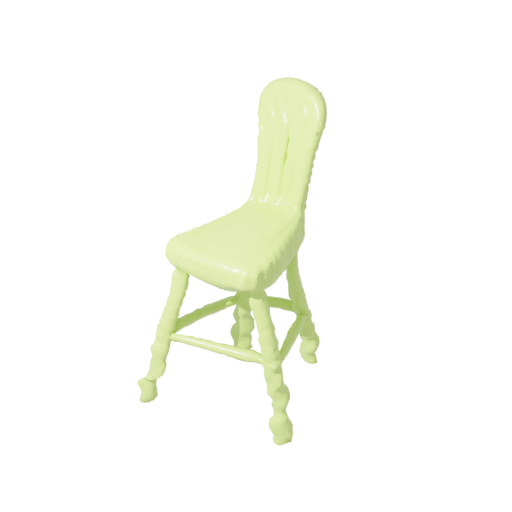} &
        \resultimg{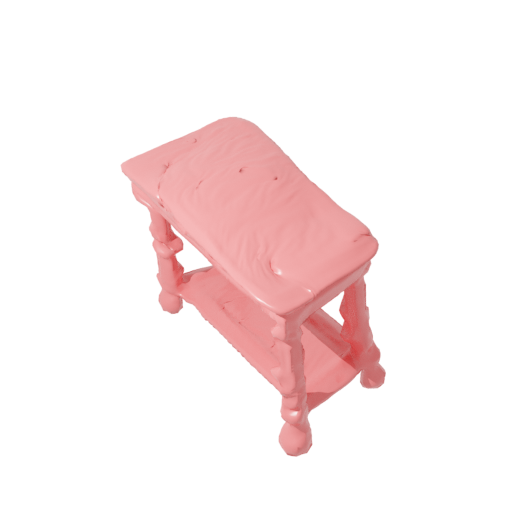} &
        \resultimg{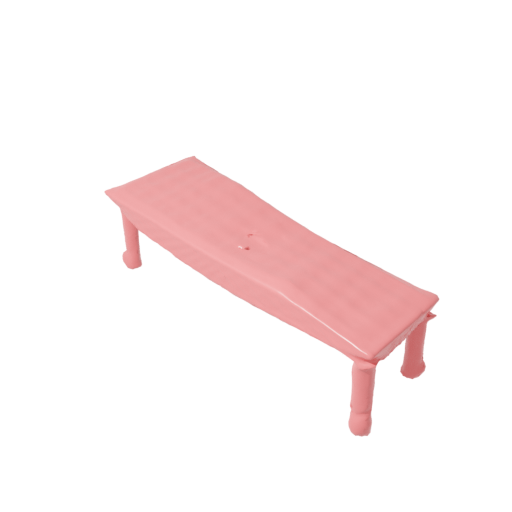} 
        \\
        \raisebox{2em}{Ours} & 
        \resultimg{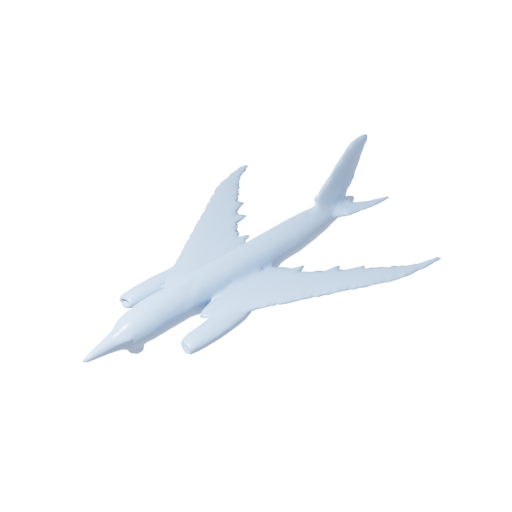} &
        \resultimg{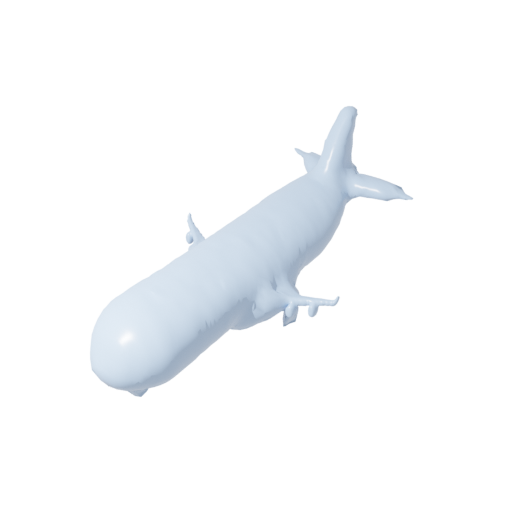} &
        \resultimg{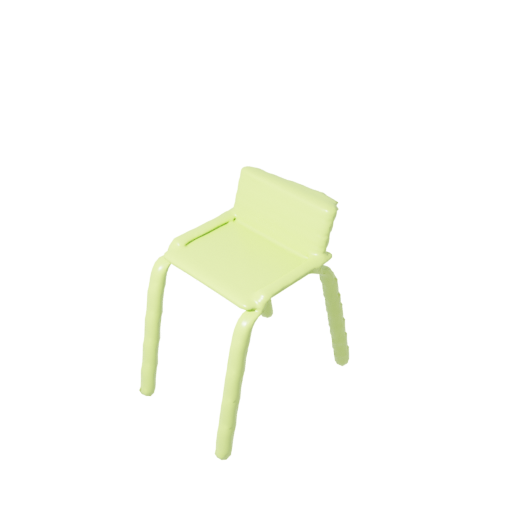} &
        \resultimg{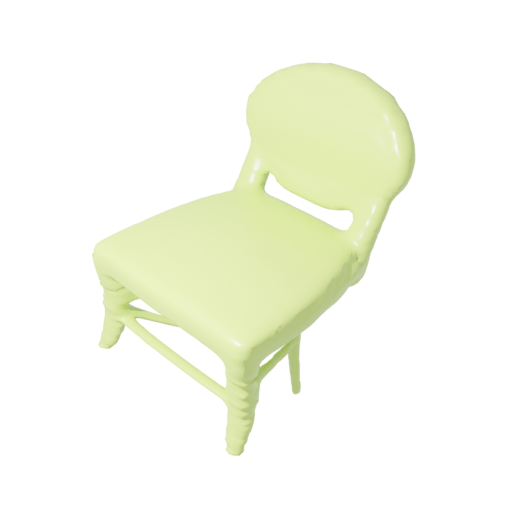} &
        \resultimg{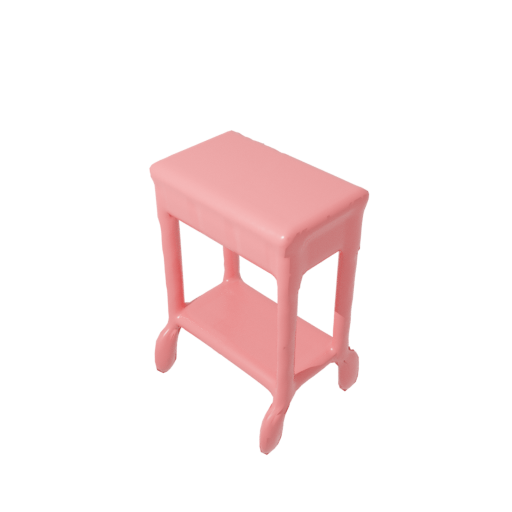} &
        \resultimg{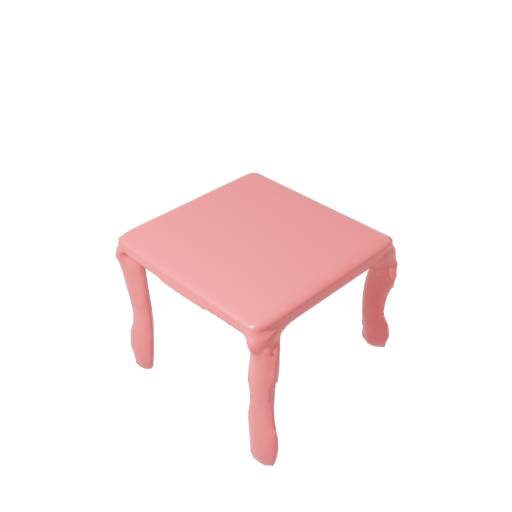} 
        \\

    \end{tabular}
    \caption{Qualitative results comparing Text2Mesh, TextDeformer, and Our method}
    \label{fig:comp}
\end{figure}

We compare the performance of our method against two baselines. The quantitative evaluation metric for the comparison are (1) The CLIP similarity (\textbf{CLIP}) score of the front-right view rendering of the deformed mesh (This evaluation view set up is different from optimization view for our method, please see figure \ref{fig:comp} as a reference for the evaluation view). (2) The averaged change of vertex Gaussian curvatures (\textbf{GC}) between the original mesh and the deformed mesh. (3) The ratio of self-intersected (\textbf{SI}) triangles of the deformed mesh. The baselines are:
\begin{enumerate}
\item{\textbf{Text2Mesh}: A method that edits a given input mesh according to a specified text prompt~\cite{text2mesh}. It jointly optimizes the vertex displacements and vertex colors. For consistency, we disabled the vertex color learning in their model and use only the vertex displacement predictions.}
\item{\textbf{TextDeformer}: A method that edits a given input mesh according to a specified text prompt~\cite{Gao_2023_SIGGRAPH}. It optimizes the Jacobians of Mesh faces to change the mesh vertex positions.}
\end{enumerate}

\begin{table}[t!]
    \centering
    \scriptsize
    \begin{tabular}{llccccccccc}
        \toprule
        \textbf{Category} & \textbf{Text} & \multicolumn{3}{c}{\textbf{Text2Mesh}}  & \multicolumn{3}{c}{\textbf{TextDeformer}} & \multicolumn{3}{c}{\textbf{Ours}} 
        \\
        \midrule
        & & CLIP $\uparrow$ & GC $\downarrow$ & SI $\downarrow$ & CLIP $\uparrow$ & GC $\downarrow$ & SI $\downarrow$ & CLIP $\uparrow$ & GC $\downarrow$ & SI $\downarrow$
        \\
        \midrule
        Airplane 
        & a fighter jet & 0.292 & 0.145 & 0.072 & 0.299 & 0.051 & 0.004 & \textbf{0.302} & \textbf{0.044} & \textbf{0.003}
        \\
        & an airship & 0.264 & 0.214 & 0.041 & 0.286 & 0.071 & 0.017 & \textbf{0.302} & \textbf{0.050} & \textbf{0.007}
        \\
        & a glider plane & 0.282 & 0.134 & 0.018 & 0.302 & 0.054 & 0.006 & \textbf{0.317} & \textbf{0.052} & \textbf{0.004}
        \\
        \midrule
        Chair 
        & a stool & 0.291 & 0.081 & 0.011 & 0.305 & 0.073 & 0.027 & \textbf{0.306} & \textbf{0.020} & \textbf{0.003}
        \\
        & a sofa & 0.301 & 0.113 & 0.029 & \textbf{0.309} & 0.076 & 0.029 & 0.299 & \textbf{0.020} & \textbf{0.003}
        \\
        & a bistro chair & 0.301 & 0.284 & 0.037 & \textbf{0.308} & 0.083 & 0.043 & 0.298 & \textbf{0.016} & \textbf{0.003}
        \\
        \midrule
        Table 
        & a dining table & 0.256 & 0.135 & 0.096 & 0.271 & 0.050 & 0.028 & \textbf{0.282} & \textbf{0.014} & \textbf{0.000}
        \\
        & a square coffee table & 0.263 & 0.094 & 0.063 & 0.285 & 0.045 & 0.016 & \textbf{0.291} & \textbf{0.014} & \textbf{0.000}
        \\
        & a nightstand & 0.256 & 0.025 & 0.002 & 0.259 & 0.049 & 0.017 & \textbf{0.272} & \textbf{0.013} & \textbf{0.000}
        \\
        \midrule
        Average 
        &  & 0.278 & 0.136 & 0.041 & 0.292 & 0.061 & 0.021 & \textbf{0.297} & \textbf{0.038} & \textbf{0.003}
        \\
        \bottomrule
    \end{tabular}
    \caption{
    Comparing Text2Mesh, TexDeformer, and Our approach on CLIP similarity, Gaussian curvature change and self-intersection ratio.
    }
    \label{tab:comparison}
\end{table}

Table \ref{tab:comparison} shows the quantitative results of the comparison (where each entry is computed as the average of $20$ randomly selected shapes from the corresponding shape category). Our framework outperforms both baselines in almost all entries across different text prompts and shape categories. Our method can produce deformed shapes with higher CLIP score, lower vertex Gaussian curvature change and lower self-intersection ratio. Please also see figure \ref{fig:comp} for some qualitative results. As you can see, our method can largely change the shape to fit the text prompt while preserving surface features of the original mesh.

\subsection{Investigation on optimization method}
We compare the performance(in terms of CLIP similarity score) of CMA-ES optimization vs gradient descent optimization on our deformation model. Please see Table~\ref{tab:optim} (where each entry is computed using $\sim5$ shapes) and Figure~\ref{fig:optim} for some results. As you can see, CMA-ES optimization outperforms gradient-based optimization in almost all cases. During our experiments, we found that gradient descent based optimization can sometimes produce good results, but CMA-ES is more reliable, robust and is easier to find a better optima. 

\begin{minipage}{\textwidth}
\begin{minipage}[b]{0.49\textwidth}
\centering

\centering
\small
\setlength{\tabcolsep}{0pt}
\renewcommand{\arraystretch}{0}
\newcommand{\resultimg}[1]{\includegraphics[trim={80pt 80pt 80pt 80pt},clip,width=0.15\linewidth]{#1}}
\begin{tabular}{lcccc}
    & \raisebox{0em}{Fighter Jet} & \raisebox{0em}{Sofa} & \raisebox{0em}{Stool} & \raisebox{0em}{Nightstand} 
    \\
    \raisebox{2em}{Original} & 
    \resultimg{figs/results/ours_fighter/4a9d28a5f272853fbbf3143b1cb6076a_originalmesh.png} &
    \resultimg{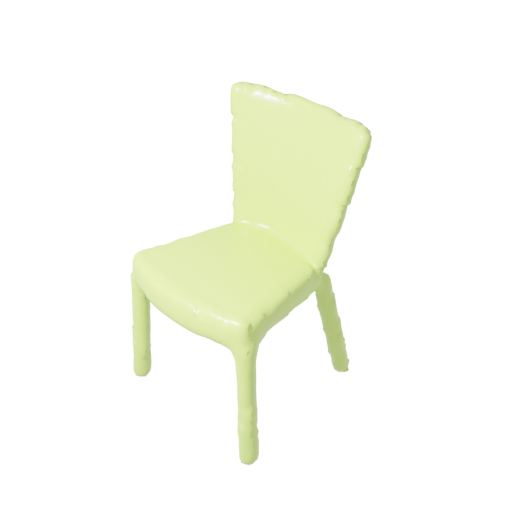} &
    \resultimg{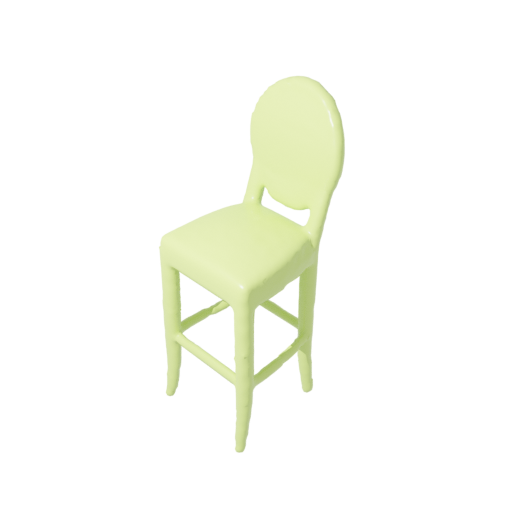} &
    \resultimg{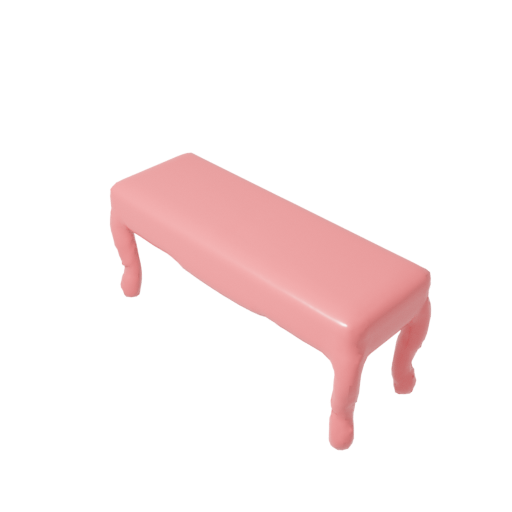} 
    \\
    \raisebox{2em}{Ours Gradient} & 
    \resultimg{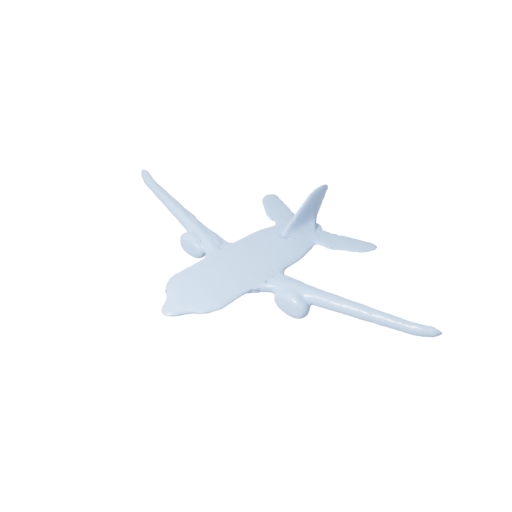} &
    \resultimg{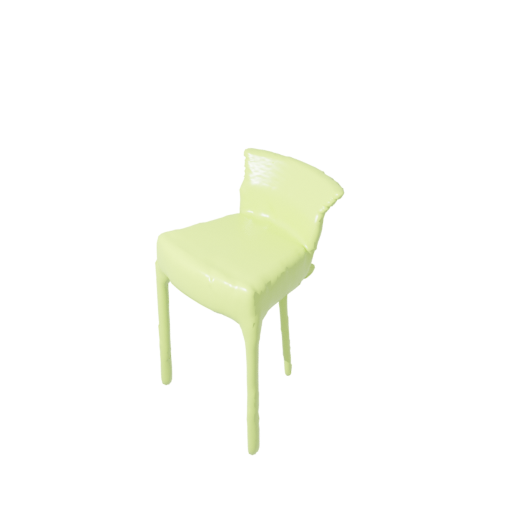} &
    \resultimg{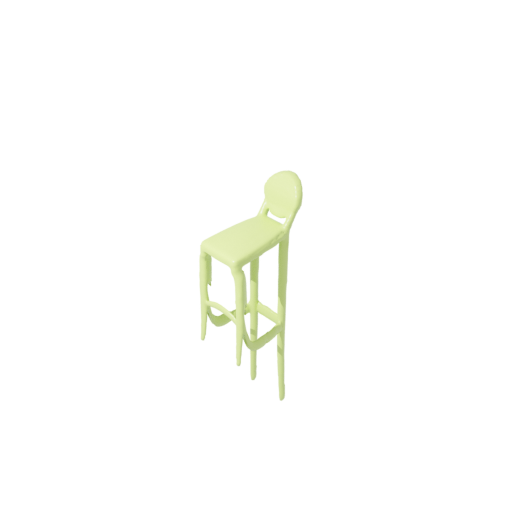} &
    \resultimg{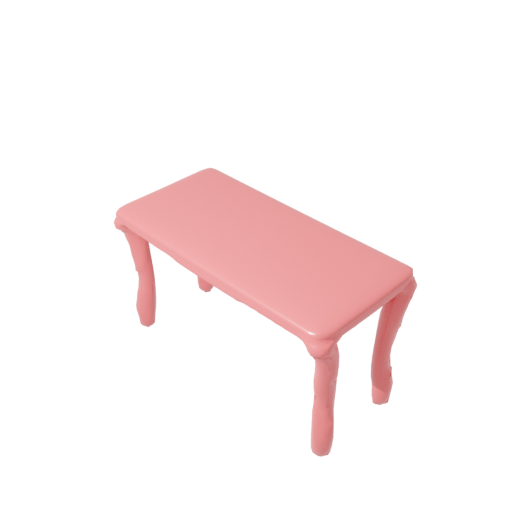} 
    \\
    \raisebox{2em}{Ours CMA-ES} & 
    \resultimg{figs/results/ours_fighter/4a9d28a5f272853fbbf3143b1cb6076a_finalmesh.png} &
    \resultimg{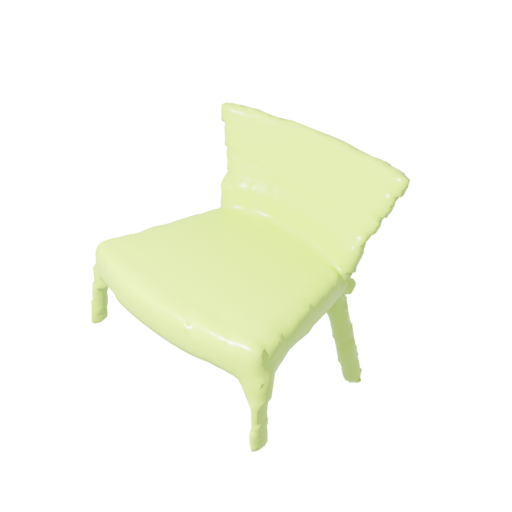} &
    \resultimg{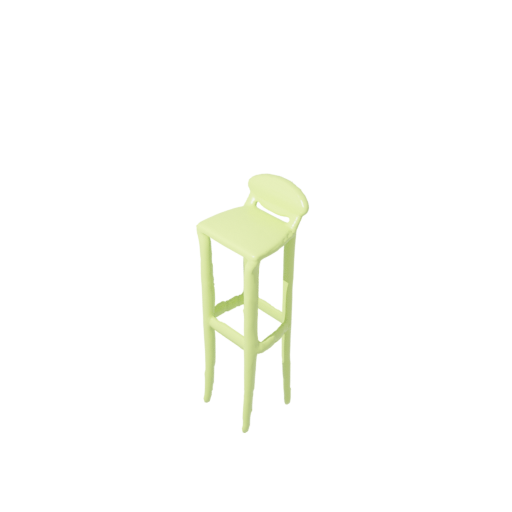} &
    \resultimg{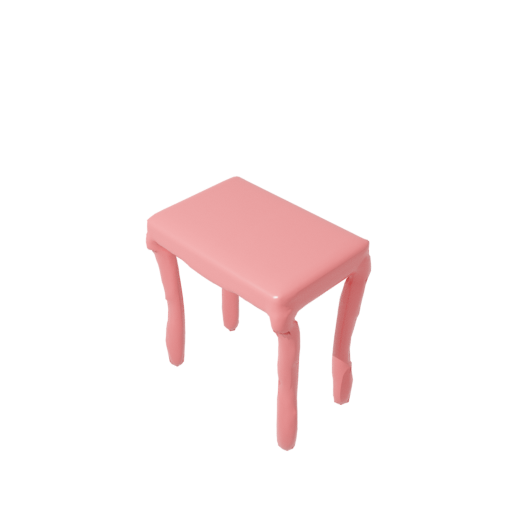} 
    \\
\end{tabular}
\label{fig:optim}
 \captionsetup{type=figure}\captionof{figure}{Qualitative results comparing gradient descent optimization and CMA-ES}
\end{minipage}
\hfill
\begin{minipage}[b]{0.49\textwidth}
 \centering
    \scriptsize
    \begin{tabular}{lcc}
        \toprule
        \textbf{Text} & \textbf{Gradient CLIP} $\uparrow$ & \textbf{CMA-ES CLIP} $\uparrow$
        \\
        \midrule 
        a fighter jet & 0.294 & \textbf{0.309}
        \\
        \midrule
        a stool & 0.281 & \textbf{0.295}
        \\
        \midrule
         a nightstand & 0.303 & \textbf{0.320}
        \\
        \midrule
        a sofa & 0.275 & \textbf{0.280}
        \\
        \midrule
        Average 
        & 0.288 & \textbf{0.301}
        \\
        \bottomrule
    \end{tabular}
    \label{tab:optim}
   \captionsetup{type=table}\captionof{table}{Quantitative results comparing gradient descent optimization and CMA-ES.}
\end{minipage}
\end{minipage}

\subsection{Additional Experiments and Limitations}
Please see the supplement for: More qualitative results; More comparison results; Ablation for scaling range; Ablation for normal consistency; Visualization of deformation sequences, Failure cases(Limitations) and more.

\section{Conclusion}
We proposed a zero-shot text-driven 3D shape deformation framework with a deformation model BoxDefGraph that deforms a given mesh to fit the given text description while still preserving the geometric details of the original mesh. Our method outperforms two strong baselines and produces visually pleasing results. Potential future work may include adding physically based constraints to the deformation and learning to generate deformers given a shape and text prompt. 

\clearpage

\bibliographystyle{plain}
\bibliography{main.bib}

\end{document}